\documentclass{article}

\usepackage[preprint]{neurips_2025}

\usepackage[utf8]{inputenc} 
\usepackage[T1]{fontenc}    
\usepackage{hyperref}       
\usepackage{url}            
\usepackage{booktabs}       
\usepackage{amsfonts}       
\usepackage{CJKutf8}        
\usepackage{graphicx}       
\usepackage{wrapfig}
\usepackage{nicefrac}       
\usepackage{microtype}      
\usepackage{xcolor}         
\usepackage{amsmath}
\usepackage{multirow}
\usepackage{multicol}
\usepackage{colortbl}
\usepackage{tcolorbox}
\usepackage{amssymb}
\usepackage{pifont}
\usepackage{subcaption}

\usepackage{natbib}

\title{ViaRL: Adaptive Temporal Grounding via Visual Iterated Amplification Reinforcement Learning}

\author{
    Ziqiang Xu$^1$,\; 
    Qi Dai$^2$,\;
    Tian Xie$^2$,\;
    Yifan Yang$^2$,\;
    Kai Qiu$^2$,\; \\
    \textbf{DongDong Chen}$^2$,\;
    \textbf{Zuxuan Wu}$^1$,\;
    \textbf{Chong Luo}$^2$ \\[2mm]
    $^1$Fudan University,
    $^2$Microsoft \\[2mm]
    \url{https://github.com/xuzq23/ViaRL}
}

\begin{document}

\maketitle

\begin{abstract}
Video understanding is inherently intention-driven-humans naturally focus on relevant frames based on their goals. Recent advancements in multimodal large language models (MLLMs) have enabled flexible query-driven reasoning; however, video-based frameworks like Video Chain-of-Thought lack direct training signals to effectively identify relevant frames. 
Current approaches often rely on heuristic methods or pseudo-label supervised annotations, which are both costly and limited in scalability across diverse scenarios. 
To overcome these challenges, we introduce ViaRL, the first framework to leverage rule-based reinforcement learning (RL) for optimizing frame selection in intention-driven video understanding.
An iterated amplification strategy is adopted to perform alternating cyclic training in the video CoT system,
where each component undergoes iterative cycles of refinement to improve its capabilities.
ViaRL utilizes the answer accuracy of a downstream model as a reward signal to train a frame selector through trial-and-error, eliminating the need for expensive annotations while closely aligning with human-like learning processes. 
Comprehensive experiments across multiple benchmarks, including VideoMME, LVBench, and MLVU, demonstrate that ViaRL consistently delivers superior temporal grounding performance and robust generalization across diverse video understanding tasks, highlighting its effectiveness and scalability. Notably, ViaRL achieves a nearly 15\% improvement on Needle QA, a subset of MLVU, which is required to search a specific needle within a long video and regarded as one of the most suitable benchmarks for evaluating temporal grounding.
\end{abstract}
\section{Introduction}

Recent advancements in OpenAI's o3 model~\citep{openai2025o3}, have demonstrated remarkable capabilities in image  understanding. The model leverages multi-turn query-based grounding and powerful reasoning abilities to process visual signals alongside textual queries. 
Inspired by this paradigm, an intriguing question arises: can video understanding be enhanced through a similar approach using temporal grounding?
While spatial grounding focuses on identifying key regions within an image, temporal grounding aims to pinpoint the most relevant frames in a video sequence.
 
Classic approaches~\citep{openai2024gpt4o,team2023gemini,bai2025qwen25vl} in video understanding often treat videos as uniform entities, processing all frames equally without considering the specific intent of the task or query. 
However, human perception is inherently intention-driven that we selectively focus on specific segments of a video based on the context or objective. 
For instance, when answering a question about a sports video, a person might pay attention to key moments of action, such as scoring or player interactions, while ignoring irrelevant segments like breaks or pauses.
This highlights the importance of temporal grounding, a process that dynamically identifies and prioritizes the most relevant frames in accordance with the given query or task. 
By aligning video processing with task-specific intent, temporal grounding enables more efficient and meaningful video understanding.

Several recent works have explored frame selection techniques and video Chain-of-Thought (CoT) pipelines. 
For instance, Hu \emph{et al.}~\citep{hu2025mllmbased} employs a learnable score query to predict importance scores for each frame, using pseudo-labels annotated by MLLMs during training. 
AKS \cite{tang2025AKS} leverages the CLIP model to compute relevance scores between each frame and the query. 
Both of them introduce extra sophisticated strategies, such as Non-Maximum Suppression (NMS) sampling, to reduce redundancy in frame selection.
CoS \cite{hu2025CoS} uses the LLaVA to evaluate whether the query elements are present in each frame to construct positive and negative shots for further co-reasoning, without temporal consideration. 
Additionally, Frame-Voyager \cite{yu2024framevoyager} applies a Direct Policy Optimization (DPO) strategy to select a group from combinations of candidate frames. 
All of these methods face limitations, specifically, these methods lack a clear training goal for assessing the quality of selected frames, making it challenging to consistently produce optimal results.

To address these limitations, we propose a novel learning framework, ViaRL, that incorporates rule-based reinforcement learning (RL) into the video CoT pipeline to optimize frame selection for temporal grounding. 
Unlike previous approach that relies on supervised fine-tuning with pseudo-labels, ViaRL uses the answer accuracy of a downstream MLLM as a reward signal, enabling a trial-and-error learning process that eliminates the need for expensive frame selection annotations. 
This approach aligns more closely with human-like learning, where individuals refine their perceptual skills through interaction and feedback rather than exhaustive supervision. 
By leveraging reinforcement learning, ViaRL dynamically trains a lightweight frame selector to identify the most relevant frames for a given query, ensuring that the model focuses on the key moments that contribute to accurate answer generation.

Inspired by the concept of iterated distillation and amplification introduced in AI 2027~\citep{Daniel2025forecast}, we adopt an iterative training strategy, referred to as Visual Iterated Amplification System, to progressively improve the performance of both the frame selector and the downstream MLLM. 
They~\cite{Daniel2025forecast,christiano2018supervising} decompose a complex problem into multiple simple sub-problems and handle them in parallel, enabling the model to progressively adapt to more challenging tasks. In comparison, we employ a sequential processing approach, which is better aligned with the requirements of our task.
Initially, the frame selector is trained using RL to optimize its selection effectiveness, based on the reward signal provided by the downstream MLLM's accuracy. 
Once the selector achieves a certain level of performance, we freeze it and fine-tune the downstream MLLM to maximize its ability to generate accurate answers using the selected frames. 
As the downstream model improves, the selector is retrained to further refine its frame selection process, creating a feedback loop where both components collectively enhance their performance. 
This iterative process ensures that the pipeline adapts to increasingly complex scenarios, enabling robust temporal grounding across diverse video understanding tasks.

By leveraging reinforcement learning and iterative optimization, ViaRL provides a scalable and human-inspired solution to intention-driven video understanding, setting a new standard for temporal grounding in multimodal tasks.
We evaluate ViaRL extensively across multiple benchmarks, including VideoMME~\cite{fu2024videomme}, LVBench~\cite{wang2024lvbench}, and MLVU~\cite{zhou2024mlvu}, demonstrating its effectiveness and scalability. 
Notably, ViaRL achieves significant improvements in temporal grounding performance compared to state-of-the-art baselines. For example, ViaRL achieves a 15\% improvement on Needle QA, a subset of MLVU that is widely regarded as one of the most suitable benchmarks for evaluating temporal grounding. 
Additionally, our experiments demonstrate that ViaRL consistently performs well across diverse visual scenarios and question types, underscoring its broad applicability and robustness.

Our contributions can be summarized as follows:
\begin{itemize}
    \item We propose ViaRL, the first framework to apply rule-based reinforcement learning to temporal grounding in video understanding tasks. By addressing the challenges posed by training signals, ViaRL establishes a learning paradigm that is both flexible and human-like.
    \item We introduce Visual Iterated Amplification Learning System, an iterative training strategy that progressively improves both frame selection and answer generation through feedback loops.
    \item Extensive experiments across multiple benchmarks featuring diverse task scenes demonstrate that ViaRL consistently outperforms strong baselines in temporal grounding tasks, highlighting its effectiveness and scalability across a broad range of task scenarios.
\end{itemize}
\section{Related Works}

\noindent\textbf{MLLMs for Video Understanding.}
Recent advancements in Multimodal Large Language Models (MLLMs) have transformed video understanding through unified representation learning, efficient temporal modeling, and scalable architectures. 
Video-LLaVA~\citep{lin2023videollava} pioneered alignment of image and video features into a shared language space, achieving state-of-the-art performance on video QA benchmarks. 
Video-LLaMA~\citep{zhang2023videollama} extended this by integrating audio-visual cues via modality-specific Q-formers. 
Scalability challenges in high-resolution and long videos were addressed by Qwen2.5-VL~\citep{bai2025qwen25vl}, which introduced dynamic FPS sampling for video processing. 
Long-context modeling saw innovations like Video-XL~\citep{shu2024videoxl}, compressing hour-long videos hierarchically, 
and LongViTU~\citep{wu2025longvitu}, emphasizing long video context and condensed reasoning. 
LLaVA-NeXT-Interleave~\citep{li2024llavanextinterleave} unifying multi-image, video, and 3D tasks. 
InternVideo2.5~\citep{wang2025internvideo2} developed compact spatiotemporal representations through adaptive hierarchical token compression.
Despite the progress, the approach still differs from the way humans process information, as it relies on uniform sampling employed by these models.

\noindent\textbf{Frame Selection.}
Efficient frame selection has become pivotal for scalable long-video understanding, evolving from traditional redundancy-reduction approaches like uniform sampling or clustering-based methods (e.g., KeyVideoLLM~\citep{liang2024keyvideollm} using text-video frame matching) to modern query-adaptive strategies. 
Early methods such as Video Summarization focused on generic keyframe extraction but lacked task-specific alignment, while contemporary techniques leverage multimodal large language models (MLLMs) for dynamic adaptation: 
M-LLM Based Frame Selection~\citep{hu2025mllmbased} employs spatial-temporal importance scoring to boost performance, 
and Frame-Voyager~\citep{yu2024framevoyager} ranks frame combinations via pre-trained Video-LLMs. 
Adaptive Keyframe Sampling (AKS)~\citep{tang2025AKS} jointly maximize prompt relevance and frame coverage through lightweight modules.
Complementary methods include Chain-of-Shot (CoS)~\citep{hu2025CoS}  exploring MLLMs’ summary capacity for binary coding and pseudo temporal grounding on long videos.

\noindent\textbf{Reinforcement Learning.}
Recent progress in RL has emphasized stable, efficient, and interpretable policy optimization. 
Trust Region Policy Optimization (TRPO)~\citep{schulman2015TRPO} introduced trust region constraints via KL divergence to ensure monotonic policy improvement, avoiding catastrophic updates in neural network training. 
Proximal Policy Optimization (PPO)~\citep{schulman2017PPO} simplified TRPO’s constraints by replacing them with a clipped objective function, enabling stable first-order optimization with lower computational costs. 
Further innovations like Group Relative Policy Optimization (GRPO)~\citep{shao2024deepseekmath} eliminated value networks in favor of group-wise KL penalties, reducing memory usage in language model alignment while maintaining training stability.
Reinforce++~\citep{hu2025reinforce++} combined REINFORCE’s simplicity with PPO-like stability mechanisms, removing critic networks to reduce complexity. 
REINFORCE Leave-One-Out (RLOO)~\citep{ahmadian2024RLOO} minimized gradient variance through leave-one-out estimation, outperforming PPO in multilingual tasks. 
Efficiency-focused methods like ReMax~\citep{li2023remax} accelerated training for large language models via greedy baselines.
Collectively, these methods bring innovations from robotic control to language alignment, emphasizing sample efficiency and stability in complex reasoning scenarios.
\section{Methods}

In this work, we propose a novel learning pipeline for video temporal grounding, referred to as Visual Iterated Amplification Reinforcement Learning (ViaRL). Our approach involves training a video frame selector that uses natural language communication to identify and convey which frames are relevant to a given query. In Section \ref{Sec: Dataset Construction}, we provide a concise overview of the training datasets used in our method. Following that, Section \ref{Sec: vision-in-the-loop Architecture} presents an overview of the architecture enabling vision-in-the-loop understanding. In Section \ref{Rule Based Reward Modeling}, we detail the rule-based rewards designed to guide the learning process. Finally, in Section \ref{Visual Iterated Amplification Reinforcement Learning}, we elaborate on the Visual Iterated Amplification Reinforcement Learning (ViaRL) framework and its implementation.

\subsection{Dataset Preparation}
\label{Sec: Dataset Construction}

We utilize a subset of the LLaVA-Video-178K~\cite{zhang2024llavavideo178k} dataset and perform a filtering operation. 
First, the CLIP-ViT-Large~\cite{radford2021clip} model is employed to select $N$ frames based on their top-$N$ cosine similarity to the question text. 
Next, a MLLM is used to predict answers based on these $N$ selected frames. 
Additionally, predictions made without incorporating visual information are considered.
Let the question be denoted as $Q$, the selected frames as $F_s$, and the correct answer as $GT$. The prediction without using visual information is denoted as $Pred_1 = MLLM(Q)$, while the prediction that incorporates the selected frames is denoted as $Pred_2 = MLLM(Q, F_s)$.

For each question-answer pair, we filter out cases where the correct prediction is made without video input, as these may have been guessed correctly. This corresponds to cases where: $GT \neq Pred_1$.
Next, we gather cases where the prediction is incorrect with visual information used, which satisfies: $GT \neq Pred_2$. 
These cases represent challenging examples that can be utilized for Reinforcement Learning (RL). 
In our implementation, Qwen2.5-VL-3B~\cite{bai2025qwen25vl} is used as the MLLM in this phrase.

\subsection{Vision-in-the-loop  Architecture}
\label{Sec: vision-in-the-loop Architecture}

\begin{figure*}[t]
\vspace{-0.7cm}
\centering
\includegraphics[width=0.99\linewidth]{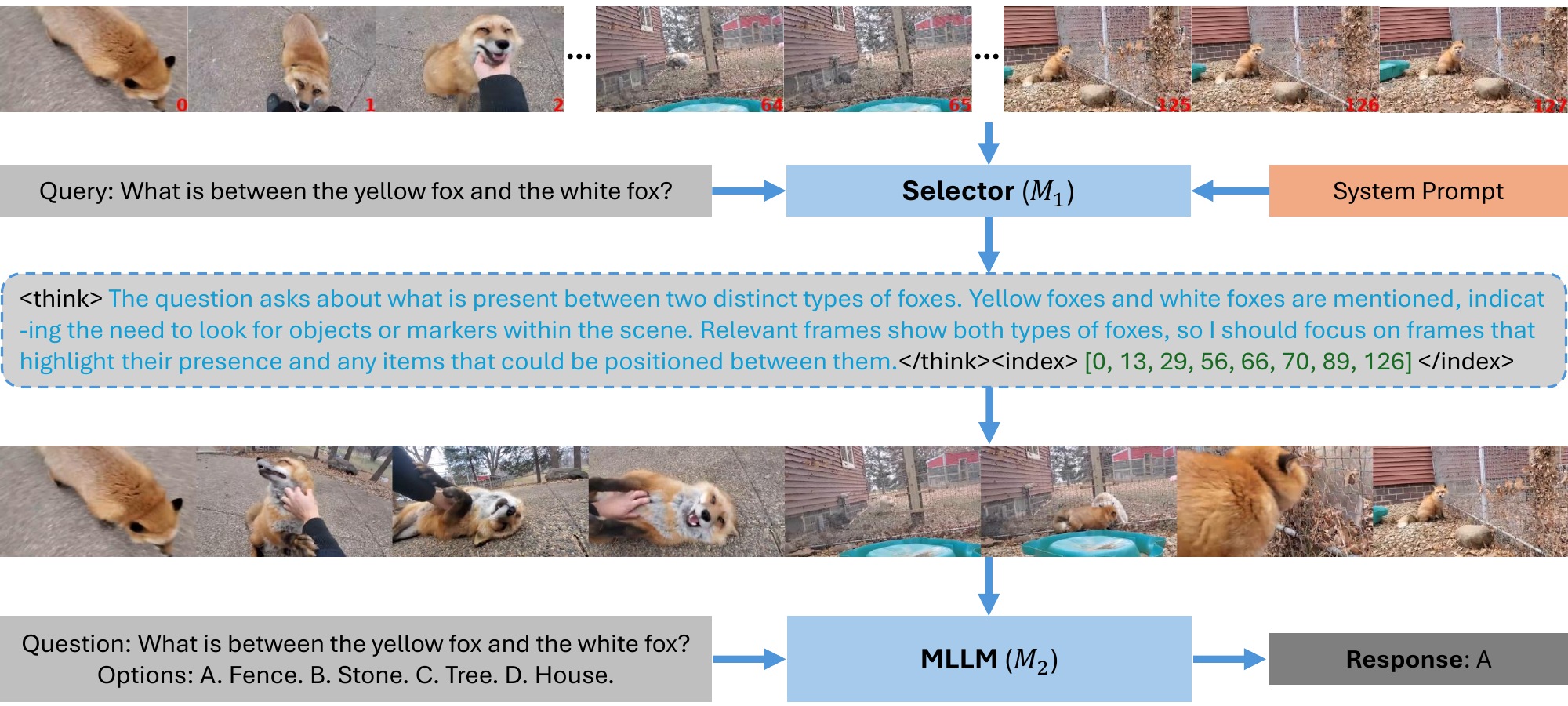}
\caption{The overall architecture of our approach. }
\label{fig:arch}
\end{figure*}

Unlike previous methods, our architecture is designed to identify the most relevant frames in response to a query text through a language-based QA approach. Similar to existing frame selection methods~\cite{hu2025mllmbased,hu2025CoS,tang2025AKS}, our architecture consists of two MLLMs. The first MLLM functions as a frame selector, while the second MLLM generates answers by thoroughly analyzing the highly relevant frames.

Previous methods, such as AKS~\cite{tang2025AKS} and CoS~\cite{hu2025CoS}, do not account for the temporal relationships between frames in their processes. Besides, Frame-Voyager~\cite{yu2024framevoyager} requires retrieving a group of frames from a vast number of combinations, which makes it computationally inefficient. While some approaches, such as those in \cite{hu2025mllmbased}, incorporate the temporal dimension, their effectiveness is limited by their reliance on pseudo labels. Additionally, both AKS~\cite{tang2025AKS} and \cite{hu2025mllmbased} utilize auxiliary selection strategies to avoid selecting redundant or overly similar frames. However, the task of selecting frames is inherently less intuitive compared to enabling a model to engage in natural language-based communication.

Our approach addresses this challenge by enabling the MLLM to directly output the serial numbers of selected video frames. To achieve this, we first address a fundamental issue: Can the model understand the serial number of each frame? Unfortunately, existing large models lack this capability. Inspired by the effective approach used in NumPro~\cite{wu2024numberit}, we directly add unique numerical identifiers to the bottom right corner of each video frame, as painted on the candidate frames in Fig. \ref{fig:arch}. This allows MLLMs to locate events temporally without requiring additional training. While NumPro uses this method to retrieve the start and end moments of events, we extend it to locate specific $N$ frames, achieving frame-level temporal grounding—a task that is significantly more challenging than clip-level grounding.

During the temporal grounding process, we further explore the reasoning capabilities of the model. First, the model analyzes the keywords in the query text to identify meaningful frames. Next, it generates detailed visual descriptions of the relevant frames to provide as much information as possible. Finally, the model outputs a list of frame indices containing $N$ selected frame numbers. As illustrated in Fig. \ref{fig:arch}, the model's reasoning process is highly informative and plays a crucial role in moment grounding.

\subsection{Rule Based Reward Modeling}
\label{Rule Based Reward Modeling}

We adopt REINFORCE++~\cite{hu2025reinforce++} as our RL algorithm, with a rule-based reward system serving as the primary training signal to effectively guide policy optimization. 
Through extensive experimentation and careful refinement of our reward design, we developed a robust rule-based reward system comprising four distinct types of rewards: Format Reward, Frame Index Reward, Answer Reward, and Response Length Reward. 
The system prompt, illustrated in \ref{prompt}, is used to guide the selector in retrieving relevant frames. For a detailed description of the system prompt, please refer to the complete version provided in the appendix.

\begin{tcolorbox}[
    colframe=teal!70!black, 
    colback=teal!10!white, 
    coltitle=white, 
    fonttitle=\bfseries, 
    title=System Prompt\label{long_open_q}, 
    sharp corners, 
    boxrule=0.5mm, 
]

You are an intelligent chatbot designed for selecting the relevant video frames according to a question.
...
Your task is to output $N_{select}$ indices of the frames that can help you answer the question better.
...
Your output should follow this format strictly:
<think> thinking about keywords and visual appearance here </think><index> target list here </index>
...
\label{prompt}
\end{tcolorbox}

\paragraph{Format Reward:}
We use regular expression extraction to enforce a structured response format. The selector is required to encapsulate its reasoning process within \texttt{<think></think>} tags and provide the target frame index list within \texttt{<index></index>} tags.
The format score (\( S_{format} \)) is computed as follows:
\begin{equation}
S_{format} =
\begin{cases}
\text{1}, & \text{if the response format is correct,} \\
\text{0}, & \text{if the response format is incorrect.}
\end{cases}
\end{equation}

\paragraph{Frame Index Reward:} 
This component evaluates the correctness of the frame indices provided in the selector’s response. To validate the indices, the model must satisfy the following conditions: the number of indices must be exactly $N$, the indices must fall within the range of valid numerical identifiers, and there must be no repetition.
The index score (\( S_{index} \)) is computed as:
\begin{equation}
S_{index} =
\begin{cases}
1, & \text{if all conditions are fully satisfied,} \\
0, & \text{if any condition is violated.}
\end{cases}
\end{equation}

\paragraph{Answer Reward:} The third component assesses the correctness of the content in the downstream model’s response. After validating the format, the model's answer is compared against the ground truth to ensure accuracy. The answer score (\( S_{answer} \)) is computed as:
\begin{equation}
S_{answer} =
\begin{cases}
2, & \text{if the answer fully matches the ground truth,} \\
0, & \text{if the answer is wrong or format/index requirements are not met.}
\end{cases}
\end{equation}

\paragraph{Response Length Reward:} 
The final component regulates the length of the content in the model’s response. 
We observe that, without proper incentives, the selector might bypass the reasoning process and directly output the index list. 
Referred from Video-R1~\cite{feng2025videor1}, we implement a length control reward to address this issue. 
This reward encourages the model to provide a detailed reasoning process alongside the index list, ensuring a more comprehensive and structured response.
The length score (\( S_{length} \)) is computed as:
\begin{equation}
S_{length} = 
\begin{cases}
0.2, & \text{if } l_{\min} \leq \text{length} \leq l_{\max}, \\
0, & \text{otherwise}.
\end{cases}
\end{equation}

According to the observation of the curve of response length varying with time, we set $l_{min} = 80$ and $l_{max} = 512$.

\subsection{Visual Iterated Amplification Reinforcement Learning}
\label{Visual Iterated Amplification Reinforcement Learning}

During the learning process, there is a critical issue that the downstream answer model can constrain the optimization of the selector. 
For instance, even when the selector chooses excellent frames, the subsequent model may produce an incorrect answer, which can severely impact the rollout selection during reinforcement learning and cause confusion for the selector. 
To address this, we propose a novel RL learning paradigm called Visual Iterated Amplification Reinforcement Learning (ViaRL).

\begin{wrapfigure}{l}{0.7\textwidth}  
    \centering  
    \includegraphics[width=\linewidth]{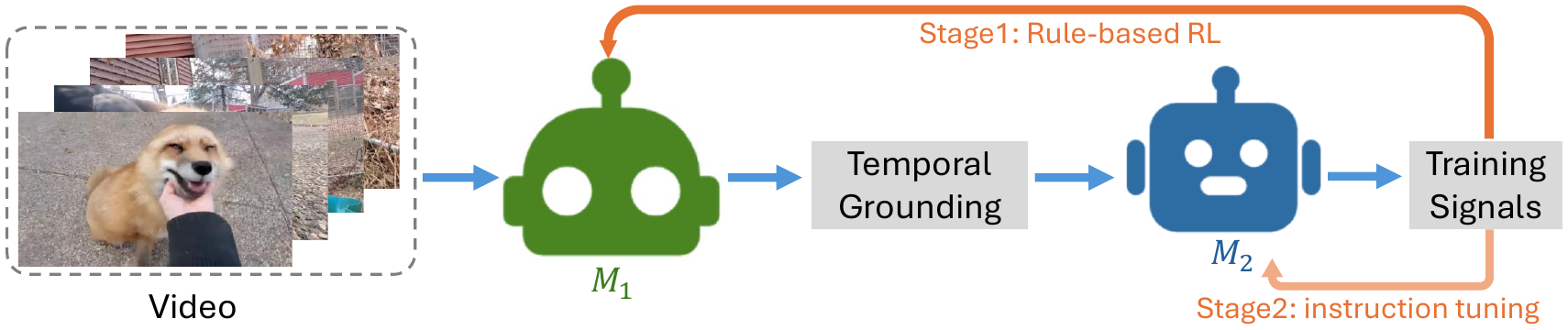} 
    \caption{Schematic of our Visual Iterated Amplification System implementation in each cycle.}
    \vspace{-12pt}
    \label{fig:ViaRL_elaboration}
\end{wrapfigure}  

In this paradigm, as elaborated in Fig \ref{fig:ViaRL_elaboration}, training is conducted in alternating phases to optimize both the selector and the answer model effectively. 
The training signals in the two stages are rewards for RL and labels for next-token prediction, respectively.
Initially, the selector undergoes reinforcement learning to achieve strong frame-picking performance. Once the selector demonstrates satisfactory results, we freeze its parameters and switch to instruction tuning of the answer model. 
As the answer model improves, we unfreeze the selector and retrain it to align with the enhanced performance of the answer model.

During the period of rule-based reinforcement learning (RL) optimization, the policy update is performed using the clipped surrogate objective and defined as follows:

\begin{equation}\small
\begin{aligned}
    \mathcal{J}_{\text{Reinforce++}}(\theta) = &\mathbb{E}_{[q \sim P(Q), \{o_i\}_{i=1}^G \sim \pi_{\theta_{\text{old}}}(O|q)]}  \\
    &~~~~~~~~~~\frac{1}{G}\sum_{i=1}^G\frac{1}{\vert o_i\vert}\sum_{t=1}^{\vert o_i\vert}\Bigg\{\min\left[\frac{\pi_{\theta}^{i,t}}{\pi_{\theta_{\text{old}}}^{i,t}}\hat{A}_{i,t}, \textrm{clip}\left(\frac{\pi_{\theta}^{i,t}}{\pi_{\theta_{\text{old}}}^{i,t}}, 1-\epsilon, 1+\epsilon\right)\hat{A}_{i,t}\right] 
    \Bigg\},
\label{equation:reinforce_plus_plus}
\end{aligned}
\end{equation}

where:
\begin{align}
\hat{A}_{i,t} = r(o_{i,<t}) - \beta \cdot \sum_{j=t}^{T} \text{KL}(j), \     \
\text{KL}(t) = \log\left(\frac{\pi_{\theta_{\text{old}}}^{i,t}}{\pi_{\text{SFT}}^{i,t}}\right).
\label{equation:adv}
\end{align}
Additionally, we normalize this advantage across the global batch for all prompts:
\begin{equation}
\hat{A}_{i,t}^{\text{norm}} = \frac{\hat{A}_{i,t}-\text{mean}\left(\hat{A}_{i,t}\right)}{\text{std}\left(\hat{A}_{i,t}\right)}.
\label{equation:batch_std_norm}
\end{equation}
The instruction tuning stage follows the general training pipeline outlined in LLaVA~\cite{liu2023llava}, which is designed to refine the model's ability to understand and respond to natural language instructions effectively.
The instruction tuning stage enhances the model's capability to handle complex queries and adapt to varying frame rates, as the frame rates after frame selection may be arbitrary. This stage ensures that the model can effectively process and reason over selected frames, regardless of their temporal distribution, while maintaining coherence and accuracy in its responses. 

This iterative and alternating training approach fosters mutual refinement between the two components, ensuring significant optimization of both models and resulting in a more robust and synergistic system.
\section{Experiments}
\label{Exp}
\subsection{Setup}

\paragraph{Benchmarks.} 
We conduct experiments on three public benchmarks to evaluate our approach. Video-MME~\cite{fu2024videomme} comprises 900 videos and 2,700 multiple-choice Question-Answer pairs, categorized into three subsets based on video duration: short (<2 minutes), medium (4$\sim$15 minutes), and long (30$\sim$60 minutes). 
MLVU~\cite{zhou2024mlvu} includes videos ranging from 3 minutes to 2 hours and spans 9 tasks, with 2,174 multiple-choice VQA pairs. 
LVBench~\cite{wang2024lvbench} features videos with an average duration of 4,101 seconds per video, which is the longest. It contains 1,549 multiple-choice VQA pairs across 6 tasks. 
Importantly, all datasets are human-annotated, ensuring high-quality labels for evaluation.

\paragraph{Training Details.} 
In this work, we utilize two models with different sizes: Qwen2.5-VL-3B as the selector and Qwen2.5-VL-7B~\cite{bai2025qwen25vl} as the answer model. 
As described in Sec. \ref{Sec: Dataset Construction}, we collect 25k pairs for reinforcement learning (RL), out of which 8k pairs are randomly selected as the final RL dataset. 
For instruction tuning, we randomly select 8k samples from the original LLaVA-Video-178k~\cite{zhang2024llavavideo178k} dataset. Across different training cycles, the same dataset is reused.

During frame selection, we choose $N$ frames from $T$ candidate frames, with the default configuration being $\{T,N\} = \{128,8\}$. The resolution of the long side for the two models is resized to $\{112,896\}$ respectively, while preserving the aspect ratio. This approach ensures that the selector model processes smaller-scale frames for efficient temporal grounding and reasoning during the thinking process.

All experiments are conducted with the selector trained using a constant learning rate of $4.0 \times 10^{-7}$, while the answer model is trained with a learning rate of $1.0 \times 10^{-6}$. The hyper-parameters for RL are set as follows: $\beta=1.0 \times 10^{-3}$, $\epsilon=0.2$.

\subsection{Performance across General Video Benchmarks}

\begin{table*}[t]
\centering
\caption{Experimental results on VideoMME (without subtitle assistance), LVBench and MLVU benchmarks. We assess the performance of ViaRL after it undergoes two cycles of learning.}
\setlength{\tabcolsep}{5pt}
\resizebox{\textwidth}{!}{
\begin{tabular}{c|c|c|cccc|c|cc}
\hline
\multirow{2}{*}{Models} & \multirow{2}{*}{Size} & \multirow{2}{*}{frames} & \multicolumn{4}{c|}{VideoMME w/o sub.} & \multicolumn{1}{c|}{LVBench val} & \multicolumn{2}{c}{MLVU Dev} \\  
\cline{4-10}  
 &  &  & Short & Medium & Long & Avg & Avg  & Needle QA & M-Avg \\ 
\hline
\multicolumn{10}{c}{Proprietary Models} \\ \hline
GPT-4o~\citep{openai2024gpt4o} & - & 384 & 80.0 & 70.3 & 65.3 & 71.9 & 30.8 & 64.8 & 64.6 \\
Gemini-1.5-Pro~\citep{team2023gemini} & - & 0.5 fps & 81.7 & 74.3 & 67.4 & 75.0 & 33.1 & - & -   \\
\hline
\multicolumn{10}{c}{Open-source MLLMs} \\
\hline
MovieChat~\cite{song2024moviechat} & 7B & 2048 & - & - & - & - & 22.5 & 24.2 & 25.8 \\
TimeChat~\cite{ren2024timechat} & 7B & 96 & - & - & - & - & 22.3 & 24.5 & 30.9 \\
VideoChat2~\cite{li2024videochat2} & 7B & 16 & 48.3 & 37.0 & 33.2 & 39.5 & - & - & 44.5 \\ %
VideoLLaVA~\cite{lin2023videollava} & 7B & 8 & 45.3 & 38.0 & 36.2 & 39.9 & - & - & 47.3 \\
Sharegpt4Video~\cite{chen2024sharegpt4video} & 7B & 16 & 48.3 & 36.3 & 35.0 & 39.9 & - & - & 46.4 \\
InternVL-V1.5~\cite{chen2024internvl15} & 20B & 10 & 60.2 & 46.4 & 45.6 & 50.7 & - & - & 50.4 \\
Video-CCAM~\cite{fei2024videoccam} & 14B & 96 & 62.2 & 50.6 & 46.7 & 53.2 & - & 73.2 & 63.1 \\
LongVA~\cite{zhang2024longva} & 7B & 128 & 61.1 & 50.4 & 46.2 & 52.6 & - & 69.3 & 56.3 \\
Video-XL~\cite{shu2024videoxl} & 7B & 128 & 64.0 & 53.2 & 49.2 & 55.5 & - & 73.8 & 64.9 \\
Kangaroo~\cite{liu2024kangaroo} & 8B & 64 & 66.1 & 55.3 & 46.7 & 56.0 & - & - & - \\
\hline
Qwen2.5-VL~\cite{bai2025qwen25vl} & 7B & 8 & 61.7 & 50.6 & 46.3 & 52.9 & 32.3 & 58.6 & 54.5 \\
Qwen2.5-VL+ViaRL & 7B & 8 & \textbf{65.1} & \textbf{56.1} & \textbf{50.8} & \textbf{57.3} & \textbf{36.9} & \textbf{73.5} & \textbf{58.2}  \\
\hline
\end{tabular}
}
\vspace{-10pt}
\label{tab:videomme}
\end{table*}

\paragraph{Temporal Grounding Analysis.} Needle QA is a subset of the MLVU benchmark that requires answering questions related to a specific segment (referred to as the needle) within a longer background video. The dataset is created by randomly inserting the needle into the background video, with a corresponding question-answer pair annotated. This sub-task best reflects the temporal grounding ability of our method. As shown in Table ~\ref{tab:videomme}, our method achieves a significant improvement, increasing from 58.6 to 73.5, which is nearly a 15\% enhancement.
Our model achieves the performance of Video-CCAM with 96 frames and Video-XL with 128 frames using only 8 frames.

\paragraph{Quantitative Analysis.}
As shown in the Table ~\ref{tab:videomme}, ViaRL brings consistent accuracy gain over three long video understanding benchmarks.
In the VideoMME benchmark, the Qwen2.5-VL+ViaRL achieves an average score of 65.1 for short videos, 56.1 for medium, 50.8 for long, and an overall average of 57.3. Compared to Qwen2.5-VL, which scores 61.7 for short, 51.9 for medium, and 47.2 for long videos with an overall average of 52.9, the Qwen2.5-VL with ViaRL model shows a marked improvement across all categories. 
In the LVBench benchmark, Qwen2.5-VL+ViaRL achieves a validation score of 36.9, outperforming other commercial models like GPT-4o and Gemini-1.5-Pro. This highlights its effectiveness in video content analysis tasks with limited frame data.
In the MLVU benchmark, particularly in the Needle QA sub-task, Qwen2.5-VL+ViaRL achieves a score of 73.5, which is a significant improvement over other models, indicating strong temporal grounding capabilities. The M-Avg score is 58.2, further showcasing its robust performance across different evaluation tasks.
Overall, the integration of ViaRL's cycle-based optimization strategy significantly enhances the video content processing capabilities of the Qwen2.5-VL+ViaRL model, making it a strong competitor against other models in scenarios with limited frames.

\paragraph{Qualitative Analysis.}

\begin{figure*}[t]
\vspace{-0.6cm}
\centering
\includegraphics[width=\textwidth]{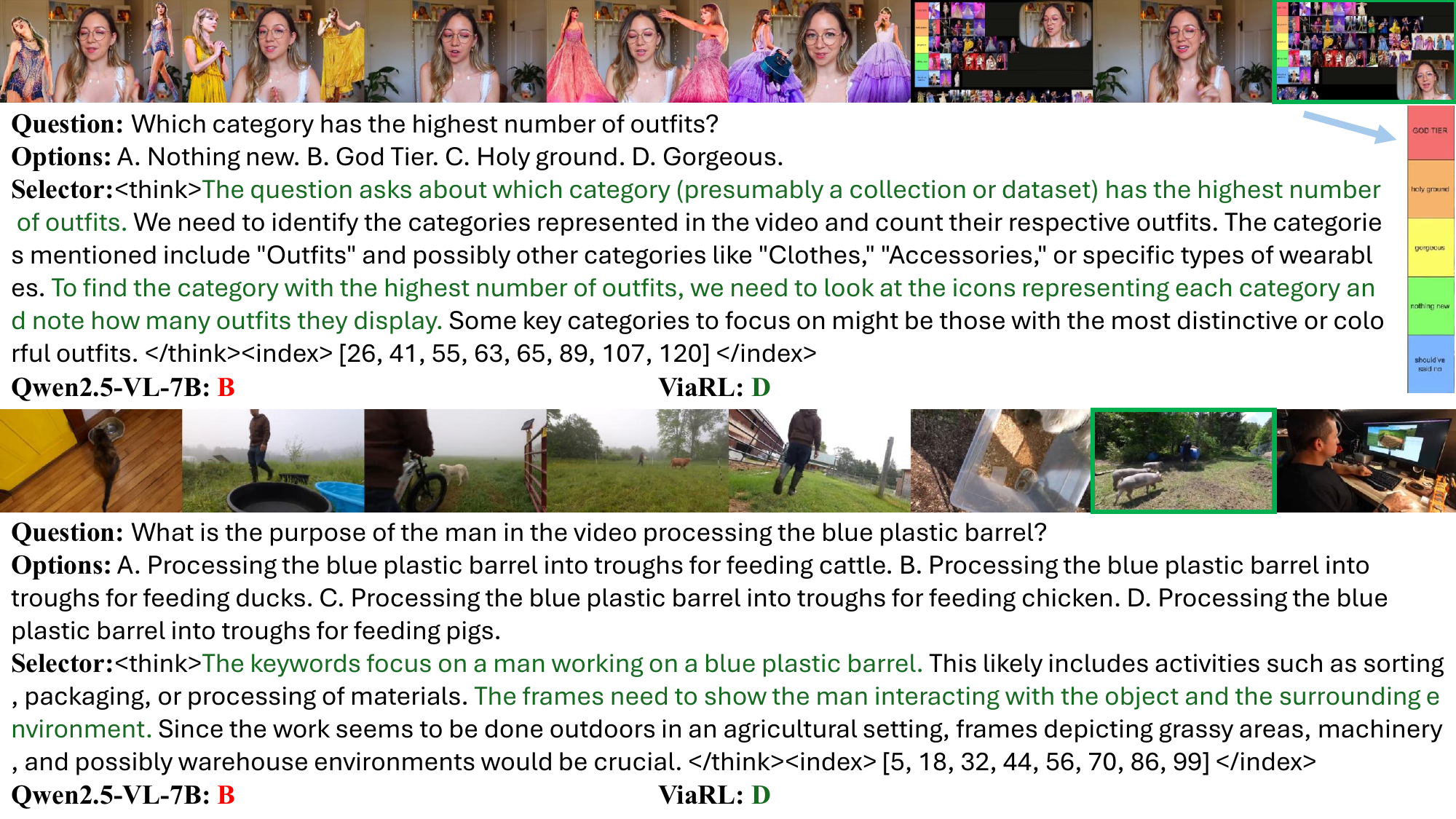}
\caption{ViaRL improves the baseline MLLMs for video understanding. The $N$ selected frames are shown. The most relevant frame is indicated by green box in each row.}
\vspace{-10pt}
\label{fig:vis_comparison}
\end{figure*}

The Fig \ref{fig:vis_comparison} showcases the visualization of frame selection process in our ViaRL, which involving visual comprehension and analysis associated with a question. 
The first case involves identifying which category has the highest number of outfits from a lineup of colorful and distinct dresses, which focuses on recognizing and counting items. 
The second case requires determining the purpose of a man interacting with a blue plastic barrel, as depicted in various scenes that suggest an agricultural or outdoor setting. It involves understanding context and potential actions based on visual cues.
We can see from the comparison of ViaRL and the baseline, ViaRL performs better by useful temporal grounding drawn by the green box.

\subsection{Ablation Study}
\label{Ablation Study}
\paragraph{Different Cycles and Stages.} 

Across these benchmarks, there is a consistent trend of performance improvement with each cycle and stage on the whole, highlighting the effectiveness of the iterative learning strategy employed by the ViaRL model. Notably, the performance of VideoMME and Needle QA on the MLVU dataset improves significantly when transitioning from cycle-stage pair $(1,2)$ to $(2,1)$, which corresponds to RL learning after completing one cycle.

As the model progresses through cycles, the rate of improvement begins to taper, indicating diminishing returns with additional cycles. 
After all, there is currently no perfect multimodal large model capable of providing ideal answers based on the selected video frames all the time. Without matching visual information, the answer will certainly be incorrect, and selecting the appropriate frames will guarantee a correct answer. 
Furthermore, the capabilities of using only 8 frames are inherently limited. 
Therefore, these limitations don't prevent us from concluding the effectiveness of multi-cycle training.

In summary, the ViaRL model demonstrates effective performance improvements through iterative cycles, with the initial cycles delivering the most substantial gains. The model's ability to progressively refine its learning across multiple stages highlights the effectiveness of a cycle-based optimization strategy in enhancing its video content processing capabilities.

\begin{figure*}[t]
\vspace{-0.8cm}
\centering
\includegraphics[width=\textwidth]{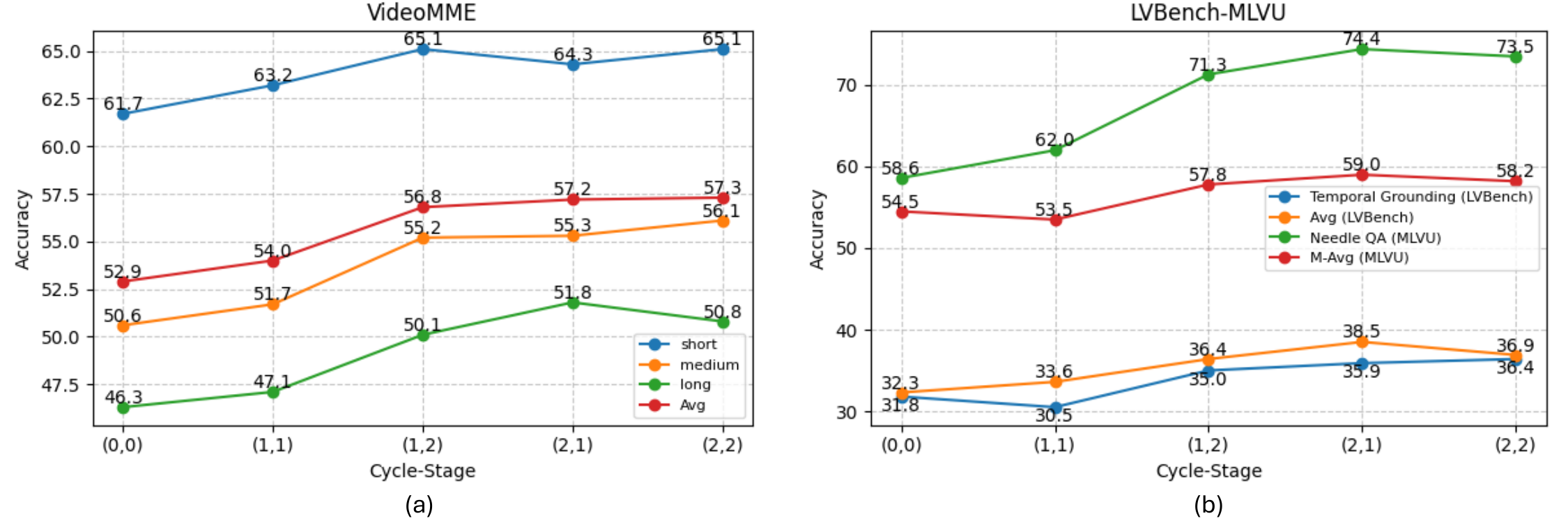}
\vspace{-0.6cm}
\caption{Performance of our ViaRL over multiple cycles and stages, attributing to the intertwined improvement of models capability during the iterative process. The horizontal axis $(i,j)$ represents the $j_{th}$ stage of the $i_{th}$ cycle. For example, $(2,1)$ indicates the evaluation model $M_1$ has learned twice, and $M_2$ has learned once. The initial state is denoted as $(0,0)$.}
\vspace{-3pt}
\label{fig:multi_cycle}
\end{figure*}

\paragraph{Different Training Recipes.} 

\begin{table}[t]  
\centering  
\resizebox{\textwidth}{!}{
\begin{tabular}{lcccccccc}  
\toprule  
{Method} & {w/ SFT} & {w/ think} & {w/ length reward} & {Data Num} & {short} & {medium} & {long} & {Avg} \\  
\midrule  
baseline & - & - & - & - & 61.7 & 50.6 & 46.3 & 52.9 \\   
SFT & \checkmark & \ding{55} & \ding{55}  & 30k & 59.9 & 49.6 & 46.2 & 51.9 \\   
RL & \checkmark & \checkmark & \checkmark & 30k+8k & 58.9 & 47.8 & 44.8 & 50.5 \\   
RL & \ding{55} &  \ding{55} & \checkmark & 8k & 58.2 & 51.3 & 45.9 & 51.8 \\  
RL & \ding{55} & \checkmark &  \ding{55}  & 8k & 59.1 & 50.7 & \textbf{47.3} & 52.4 \\
RL & \ding{55} & \checkmark & \checkmark & 8k & \textbf{63.2} & \textbf{51.7} & 47.1 & \textbf{54.0} \\   
\bottomrule  
\end{tabular}  }
\caption{VideoMME w/o sub. performance metrics for different training recipes. All RL evaluations are conducted after the $1_{th}$ stage of the $1_{th}$ cycle.}  
\label{tab:video_mme_performance} 
\vspace{-24pt}
\end{table}

Given that the initial model exhibits poor frame-level temporal grounding capacity with text queries, we incorporate the model fine-tuned through Supervised Fine-Tuning (SFT) as a starting point for reinforcement learning (RL). 
As outlined in Sec \ref{Sec: Dataset Construction}, we construct the SFT dataset by collecting cases where the correct answer is predicted using the selected frames, satisfying the conditions: $GT \neq Pred_1$ and $GT = Pred_2$. These conditions indicate that the selected $N$ frames are relevant and can serve as pseudo labels for the SFT process.

We have collected 30k SFT data points, each paired with pseudo-labels specifying $N$ relevant frame indices. 
Leveraging this dataset, we use Supervised Fine-Tuning (SFT) to train a model capable of directly predicting the relevant frames.
Subsequently, we use this SFT-trained model as the initial policy for reinforcement learning (RL), leveraging its existing ability for frame grounding to some extent. 
However, the results deteriorate, with the average accuracy decreasing from 51.9\% to 50.5\%. 
Based on our observations, this decline is primarily due to the SFT-trained model inheriting the shortcomings of the CLIP model. For example, consider a query like "What is the logo displayed at the beginning of the video?". The model struggles to interpret the temporal word "beginning" and instead focuses disproportionately on the entity "logo". As a result, it might select $N$ frames predominantly from the end of the video, where the "logo" appears with higher confidence, rather than correctly grounding the query to the initial frames. Therefore, the SFT-trained model is not an ideal starting point for RL.

By leveraging the training signals provided by the answer model, the "RL without thinking" approach can also be trained. However, its average accuracy is 51.8\%, and the average accuracy without applying the length reward is 52.4\%. Both are noticeably lower compared to our method, which achieves an average accuracy of 54.0\%. 
These results indicate that our approach, which incorporates a rich and meaningful thinking process, significantly enhances the model's ability to select relevant frames more effectively.
\section{Conclusion}

In this work, we proposed ViaRL, a novel framework that integrates rule-based reinforcement learning into the video CoT pipeline to address the challenges of temporal grounding in multimodal video understanding tasks. By leveraging a reward-driven trial-and-error learning process inspired by human-like perceptual refinement, ViaRL eliminates the lacking of frame selection annotations and dynamically trains a lightweight frame selector to focus on the most relevant frames for further accurate answer generation. Through an iterative optimization strategy, referred to as the Visual Iterated Amplification Learning System, ViaRL progressively enhances the performance of both the frame selector and downstream multimodal large language models (MLLMs), adapting to increasingly complex scenarios and ensuring robust temporal grounding.

\newpage
\bibliography{references}{}
\bibliographystyle{plain}


\appendix
\newpage
\section{Details of System Prompt}
\begin{tcolorbox}[
    colframe=teal!70!black, 
    colback=teal!10!white, 
    coltitle=white, 
    fonttitle=\bfseries, 
    title=System Prompt, 
    sharp corners, 
    boxrule=0.5mm, 
]
You are an intelligent chatbot designed for selecting the relevant video frames according to a question.

User will provide you a video with $N_{candidate}$ frames and a short question.

The red numbers in the bottom right corner of each frame represent the frame indice. The frame index is an integer in the range of 0 to $N_{candidate}-1$.

Your task is to output $N_{select}$ indices of the frames that can help you answer the question better.

Here's how you can accomplish the task:

1. Think about the keywords from the question:

- Check if the physical entities are mentioned.

- Check if the occurrence time is mentioned.

- Check if the place or location is mentioned.

- Check if the action is mentioned.

2. Provide the appearance reference based on the keywords and video:

- Describe the visual appearance of the $N_{select}$ frames that are most relevant to the keywords.

3. Provide the target list: 

- A list of $N_{select}$ frame indices, that the corresponding frames are most helpful to answer the question.

Your output should follow this format strictly:

<think> thinking about keywords and visual appearance here </think><index> target list here </index>.

Specific requirements are as follows:

**Ensure that anyone can uniquely identify these target frames in the video through the references.**

**Ensure that the references are complete and independent.**

**Don't output the words '<think> thinking about keywords and visual appearance here </think>' directly.**

**Ensure that the list consists of $N_{select}$ values.**

\end{tcolorbox}

\section{More Visualization Results}
\begin{figure*}[t]
\centering
\includegraphics[width=\textwidth]{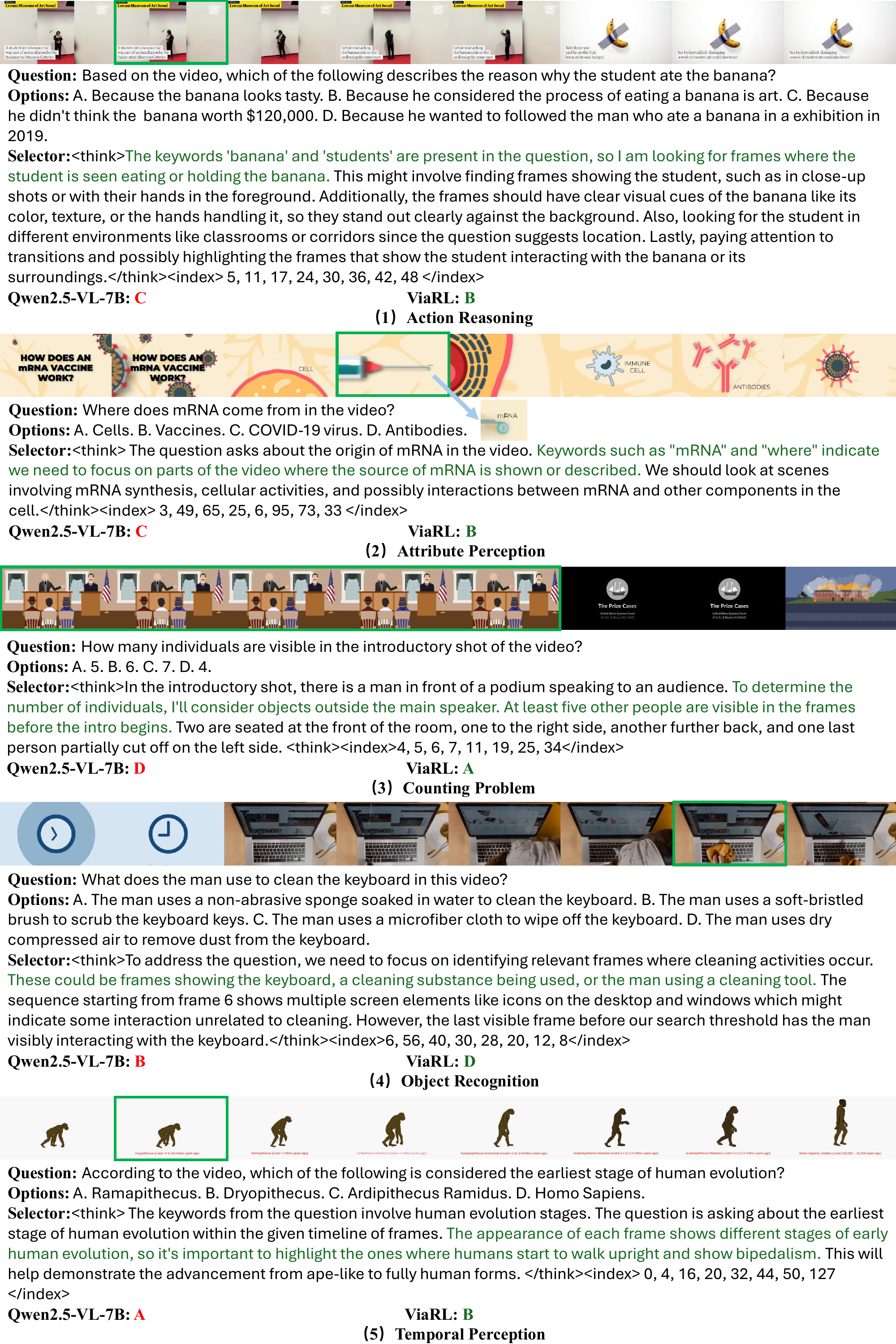}
\caption{Visualization across diverse scenarios on VideoMME.}
\label{fig:more_vis_videomme}
\end{figure*}

\begin{figure*}[t]
\centering
\includegraphics[width=\textwidth]{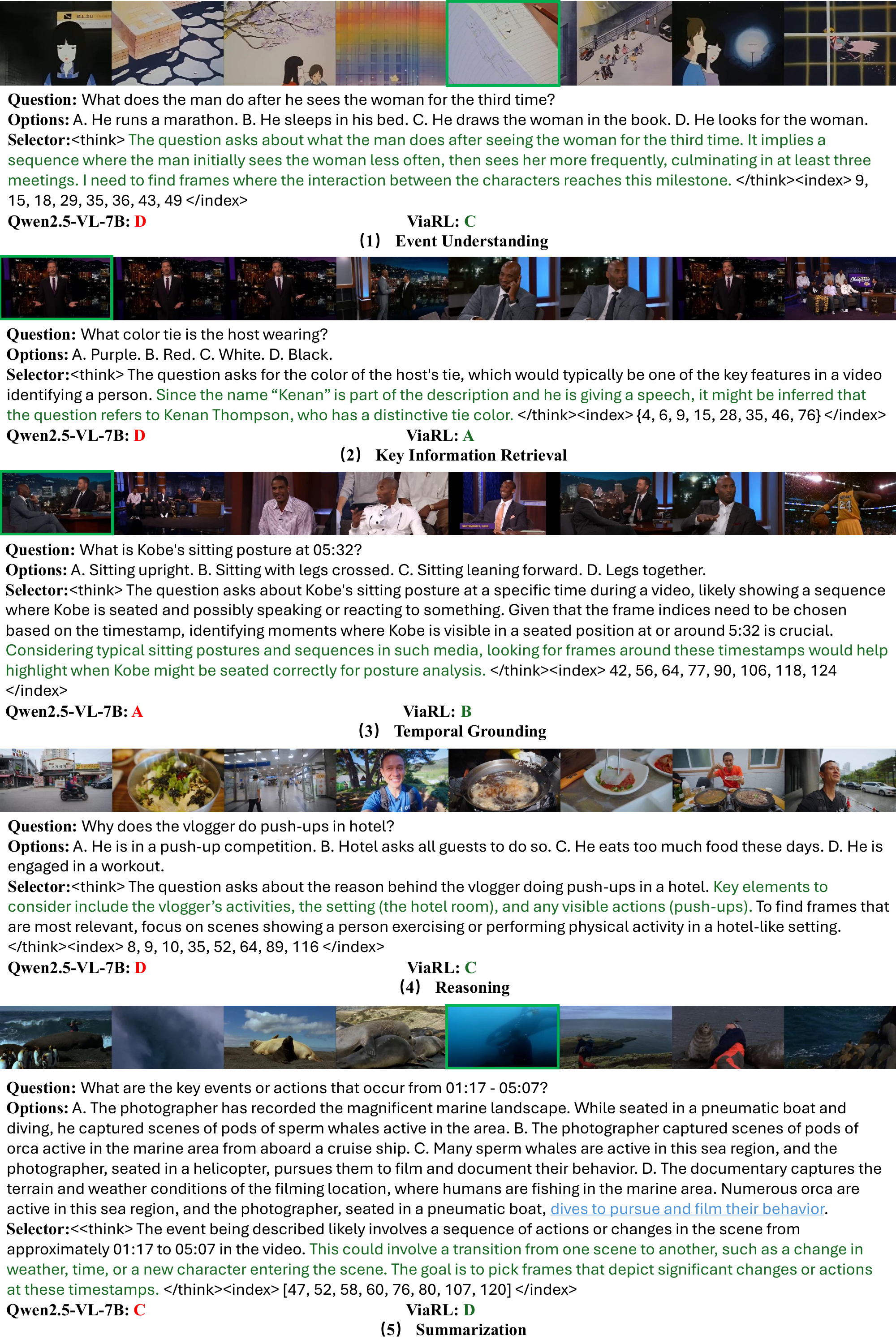}
\caption{Visualization across diverse scenarios on LVBench.}
\label{fig:more_vis_lvbench}
\end{figure*}

\begin{figure*}[t]
\centering
\includegraphics[width=\textwidth]{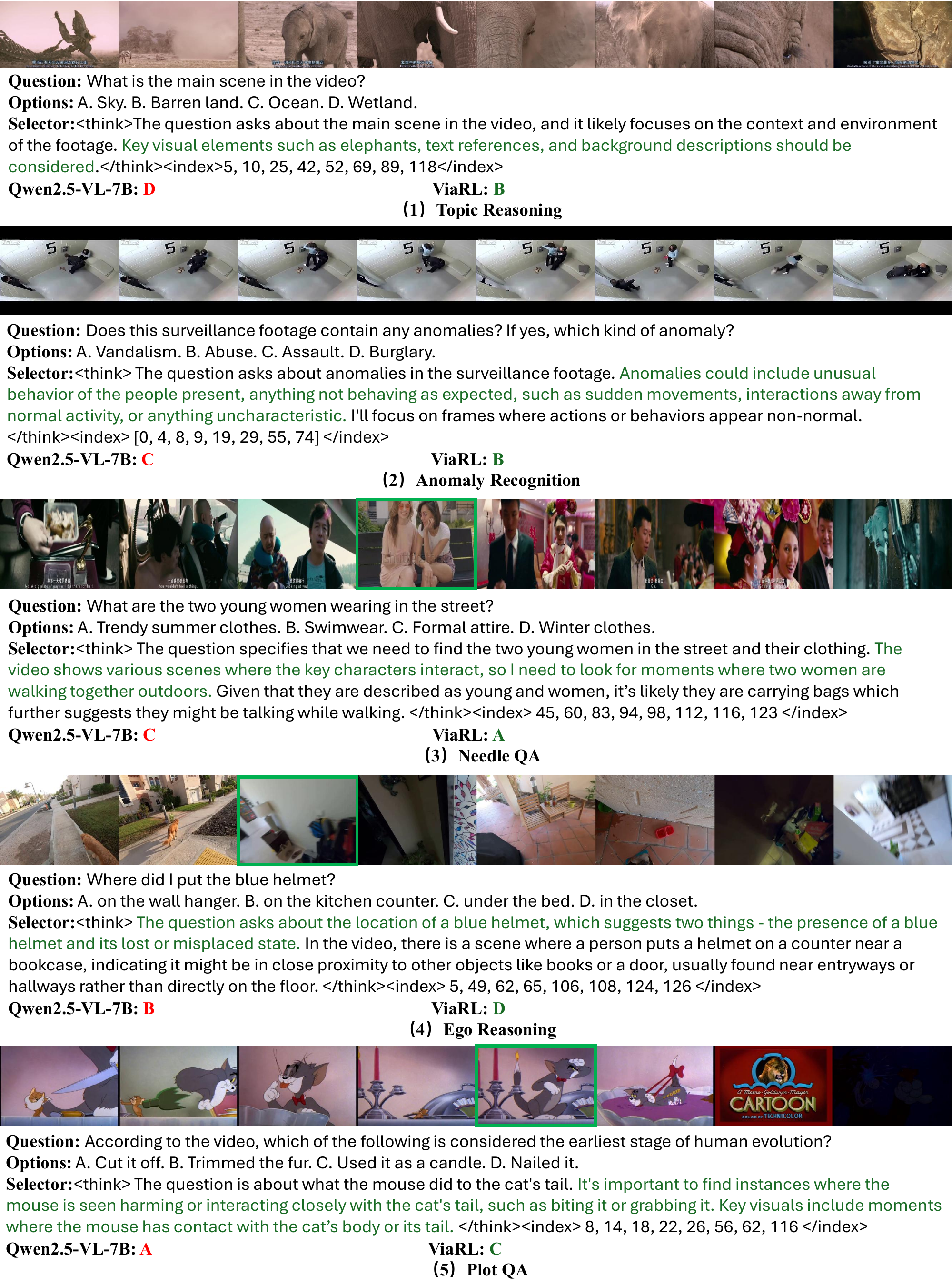}
\caption{Visualization across diverse scenarios on MLVU.}
\label{fig:more_vis_mlvu}
\end{figure*}

In Figures \ref{fig:more_vis_videomme}, \ref{fig:more_vis_lvbench}, and \ref{fig:more_vis_mlvu}, we present additional examples of video understanding results achieved by our ViaRL. These examples demonstrate the consistent performance of ViaRL across a wide range of visual scenarios and question types, highlighting its versatility and robustness in diverse applications.

\end{document}